%% file: main.tex
\documentclass[lettersize,journal]{IEEEtran}
\usepackage{amsmath,amsfonts}
\usepackage{algorithmic}
\usepackage{algorithm}
\usepackage{array}
\usepackage[caption=false,font=normalsize,labelfont=sf,textfont=sf]{subfig}
\usepackage{textcomp}
\usepackage{stfloats}
\usepackage{url}
\usepackage{verbatim}
\usepackage{graphicx}
\usepackage{cite}

\usepackage{algorithm,algorithmic}
\usepackage{float}
\usepackage{multirow}
\usepackage{float}
\usepackage{color} 
\usepackage{comment}
\usepackage{amsthm}
\usepackage{xspace}

\newcommand{\ours}[0]{SimuLine\xspace}
\newcommand{\etal}{\textit{et al}.}

\newcounter{tightlistcnt}
\newenvironment{tightitemize}{%
	\begin{list}{\textbullet}{\usecounter{tightlistcnt}%
			\topsep 0in
			\partopsep 0in
			\itemsep .2em
			\parsep 0in
			\leftmargin 0em
			\rightmargin 0in
			\listparindent 0em
			\itemindent .75em
			\labelsep .75em
			\labelwidth 0in
		}%
	}{%
	\end{list}%
}

\begin{document}

\title{Simulating News Recommendation Ecosystem for Fun and Profit}

\author{Guangping~Zhang,
        Dongsheng~Li,
        Hansu~Gu,
        Tun~Lu,
        Li~Shang,
        and~Ning~Gu
\thanks{Manuscript received April 19, 2021; revised August 16, 2021. This work was supported by the National Natural Science Foundation of China (NSFC) under Grants 62172106 and 61932007. (Corresponding authors: Hansu Gu and Tun Lu.)}
\thanks{Guangping Zhang, Tun Lu, Li Shang, and Ning Gu are with School of Computer Science, and with Shanghai Key Laboratory of Data Science, Fudan University, Shanghai, 201203, P. R. China (e-mail: gpzhang20@fudan.edu.cn; lutun@fudan.edu.cn; lishang@fudan.edu.cn; ninggu@fudan.edu.cn). }
\thanks{Dongsheng Li is with Microsoft Research Asia, Shanghai, and an adjunct professor with School of Computer Science, Fudan University, Shanghai, 201203, P. R. China (e-mail: dongshengli@fudan.edu.cn).}
\thanks{Hansu Gu is with Amzon.com, Seattle, USA (e-mail: hansug@acm.org).}
}

\markboth{Journal of \LaTeX\ Class Files,~Vol.~14, No.~8, August~2021}%
{Shell \MakeLowercase{\textit{et al.}}: A Sample Article Using IEEEtran.cls for IEEE Journals}

\IEEEpubid{0000--0000/00\$00.00~\copyright~2021 IEEE}

\maketitle

\begin{abstract}
Understanding the evolution of online news communities is essential for designing more effective news recommender systems. However, due to the lack of appropriate datasets and platforms, the existing literature is limited in understanding the impact of recommender systems on this evolutionary process and the underlying mechanisms, resulting in sub-optimal system designs that may affect long-term utilities. 
In this work, we propose \ours, a simulation platform to dissect the evolution of news recommendation ecosystems and present a detailed analysis of the evolutionary process and underlying mechanisms. 
\ours first constructs a latent space well reflecting the human behaviors, and then simulates the news recommendation ecosystem via agent-based modeling. 
Based on extensive simulation experiments and the comprehensive analysis framework consisting of quantitative metrics, visualization, and textual explanations, we analyze the characteristics of each evolutionary phase from the perspective of life-cycle theory, and propose a relationship graph illustrating the key factors and affecting mechanisms. 
Furthermore, we explore the impacts of recommender system designing strategies, including the utilization of cold-start news, breaking news, and promotion, on the evolutionary process, which shed new light on the design of recommender systems. 
\end{abstract}

\begin{IEEEkeywords}
News recommendation, agent-based modeling, simulation, online community, life-cycle, social impact.
\end{IEEEkeywords}

\input{intro}

\input{related}

\input{method}

\input{experiments}

\section{Conclusion and Future Work}
\label{sec:conclusion}

This paper presents \ours, a simulation platform to dissect the evolution of news recommendation ecosystems, and provides a detailed analysis of the evolutionary process. 
\ours constructs a latent space well reflecting the human behaviors, based on which we simulate the NREs via agent-based modeling. 
\ours dissects the lifecycle of the NRE evolution, which consists of the start-up, growth, and maturity\&decline phases. 
We analyze the characteristics of each phase and propose a relationship graph illustrating the key factors and affecting mechanisms. 
In the end, we explore the impacts of recommender system designing strategies, including the utilization of cold-start news, breaking news, and promotion, on the evolutionary process. 

In the future, we will consider supporting the textual content generation of the synthetic news and the action modeling of social network activities for more powerful simulation. 
Besides, \ours also benefits the evaluation of recommendation algorithms.
In particular, some debiasing recommendation algorithms~\cite{chen2020bias} are recently proposed aiming at handling the exposure bias, which is the direct cause of user differentiation and topic convergence. 
Since this paper focuses on the systematic design of recommender systems rather than on specific recommendation algorithms, we leave this issue as an open topic and hope that \ours could facilitate future research in this direction.

\bibliographystyle{IEEEtran}
\bibliography{TCSS}

\vfill

\end{document}

%% file: intro.tex
\section{Introduction}
\label{sec:introduction}

\IEEEPARstart{D}{ue} to the proliferation of social media, people are increasingly relying on online news communities to publish and acquire news. 
There are millions of news posted to online news communities by content creators and read by a large number of users with the distribution of recommender systems~\cite{li2019survey,wu2022personalized}. 
Online news communities are continuously and dynamically evolving with the production and consumption of news content~\cite{merchant2014click, baker2017rational, liu2021interaction}. 
Like other online communities, online news communities evolve in line with the famous life-cycle theory~\cite{mueller1972life}, i.e., they would go through the phases of ``start-up''-``growth''-``maturity''-``decline'' in turn.
Through the lens of life-cycle theory, extensive works have investigated the evolving patterns of online communities and given suggestions for operations at each stage~\cite{iriberri2009life}. 
However, the impact of recommender systems, one of the most important technical infrastructures, on the lifecycle of online news communities is yet unclear.
The life-cycle theory is critical for analyzing the long-term utility of recommender systems and suggesting optimal designs.
Therefore, this paper will focus on the following three research questions and try to answer them via simulation experiments:
1) What are the characteristics of each phase of the evolutionary lifecycle of recommender system-driven online news communities (news recommendation ecosystems or NREs for short)?
2) What are the key factors driving the evolution of NREs and how do these factors affect the evolutionary process?
3) How can we achieve better long-term multi-stakeholder utilities and avoid communities from falling into ``decline'' through the design strategy of recommender systems?

Some attempts have been made to study the social impact of recommender systems through simulations~\cite{stavinova2022synthetic,luo2022mindsim}.
These works seek to explain phenomenons, such as filter bubble~\cite{geschke2019triple, jiang2019degenerate, aridor2020deconstructing}, popularity bias~\cite{sun2019debiasing, zhao2021popularity} and content quality~\cite{ciampaglia2018algorithmic}. 
However, as these works focus on some specific issues, they ignore the following limitations when designing their simulator, resulting in the difficulty in modeling the evolutionary process of NREs: 
1) The exposure bias of the original datasets is not handled properly. The synthetic data generation for the simulators needs to be based on real-world datasets to ensure authenticity and reliability. As the original datasets are collected from users already exposed to algorithmic recommendations, the exposure bias is introduced unavoidably, resulting in troubles such as distribution distortion and algorithm confounding~\cite{chaney2018algorithmic}. 
2) The latent space~\footnote{In this paper, we use ``latent space'' to refer to the real user interest space modeled by the simulators, and ``embedding space'' to refer to the user interest learned by the recommendation algorithm within the simulation system.\\ \\} could not well reflect the human behaviors due to excessive idealization or limited explainability. Some works regard each dimension of the latent space as a disentangled attribute and sample the latent vectors from pre-specified distributions~\cite{krauth2020offline, lucherini2021t}. This approach is over-idealistic and ignores the correlation between different features as well as their different importance. The other works learn the latent vectors from historical interactions using matrix factorization algorithms~\cite{bountouridis2019siren, yao2021measuring, zhou2021longitudinal}. However, the learned latent space is abstractive and less explainable, restricting the understanding of the evolutionary process. 
3) Existing frameworks are not authentic enough to reflect real-world scenarios. On one hand, existing frameworks confuse the concepts of recommendation algorithms and recommender systems. In addition to algorithmic recommendations, recommender systems need to address the cold start issues, and strategic recommendations such as utilizing trending news and promotions should also be considered. On the other hand, existing frameworks ignore some key factors in user behavior modeling, such as the distinction between pre-consumption and post-consumption behaviors, content quality modeling, etc~\cite{ma2018entire}.

In this paper, we first propose \ours, an advanced simulation platform for dissecting the evolution of NREs. \ours first performs synthetic data generation based on real-world datasets. To address the inherent exposure bias of the original datasets, \ours introduces the Inverse Propensity Score (IPS) for bias elimination~\cite{saito2020unbiased}. In order to build a latent space close to the human decision process, we introduce the pre-trained language models (PLMs)~\cite{pennington2014glove, devlin2018bert}, which are trained on large-scale corpus and proved to exhibit certain rules of human cognition~\cite{allen2019analogies, ethayarajh2019towards}. Then we simulate the interaction behavior of users, content creators, and recommendation systems through Agent-based Modeling.
Based on the extensive analysis framework, this paper attempts to answer the above three research questions with life-cycle analysis, key factors and affecting mechanisms analysis, and recommender system designing strategies experiments.

Our major contributions are summarized as follows:
\begin{itemize}
    \item We propose \ours, a novel news recommendation ecosystem simulation platform supporting effective and reliable simulation of users, content creators, and recommender systems. Utilizing the PLMs and IPS, \ours operates in a realistic and explainable latent space, which is consistent with human behaviors and can get rid of the inherent exposure bias of the original datasets. 
    \item \ours is the first work to dissect the life-cycle of the news recommendation ecosystem via simulation. 
    We attempt to understand the characteristics of each phase, and propose a relationship graph illustrating the key factors and affecting mechanisms, shedding new light on designing recommender systems responsible for long-term utilities. 
    \item The simulation platform and simulation experiments will be opensourced~\footnote{Code will be publicly available upon acceptance of this paper.}, further contributing to the community for understanding and analyzing the news recommendation ecosystems in the future. 
\end{itemize}

%% file: related.tex
\section{Related Work}
\label{sec:related_work}

Considering the high cost of online experiments and the low flexibility of dataset-based offline experiments, simulators have been widely leveraged in building recommendation models and understanding recommender systems. 
Earlier simulators are designed for generating synthetic data to overcome the limited availability of appropriate datasets~\cite{del2017datagencars, jakomin2018generating, slokom2018comparing}. 
These works learn the distributions of the incomplete original datasets, and then apply data generating schemes for specific scenarios, such as context-awareness~\cite{del2017datagencars}, streaming~\cite{jakomin2018generating} and privacy preservation~\cite{slokom2018comparing}. 

More recent simulators could be divided into two main categories according to their purposes.
The first category contains simulators built for reinforcement learning tasks~\cite{rohde2018recogym, chen2019generative, ie2019recsim, mladenov2021recsim, shi2019virtual, huang2020keeping, mcinerney2021accordion, zhao2021usersim, wang2021learning, luo2022mindsim}, which are designed for optimizing the long-term user utilities in multi-round interactions. 
Regarding the recommender system as an agent, these works construct user simulators, which could continuously respond to the recommendations, as the environment.
Besides, some methods pay additional attention to content provider modeling~\cite{mladenov2021recsim}, visit pattern modeling~\cite{mcinerney2021accordion}, debiasing\cite{huang2020keeping}, etc., in order to make the simulated users more realistic.
Most of these approaches are implemented with probabilistic programming to support model optimization, which demands the probability of the interaction process tractable. But this requirement also introduces limitations on simulation scale and agent flexibility.

The second category contains simulators designed for understanding the social impact of recommender systems in long-term interactions. 
Compared to the simulators designed for reinforcement learning, these simulators loosen the requirement for probabilistic tracking and pay more attention to building complex and realistic simulation systems. 
These works firstly propose the well-designed simulation frameworks, reproduce the concerned phenomenons and then explain the phenomenons using the simulating data. 
Following this methodology, existing literature analyses the role of recommender systems in 
filter bubble~\cite{jiang2019degenerate}, 
echo chamber~\cite{geschke2019triple, jiang2019degenerate}, 
popularity bias~\cite{yao2021measuring}, 
content quality~\cite{ciampaglia2018algorithmic}, 
feedback loops~\cite{zhou2021longitudinal}, 
user homogenization~\cite{chaney2018algorithmic, aridor2020deconstructing, lucherini2021t}, 
algorithmic recommendation coverage~\cite{bountouridis2019siren} and 
exploration-exploitation trade-off~\cite{krauth2020offline}. 
\ours further extends this line of work via improved synthetic data generation and simulation designs, enabling more comprehensive analysis.

%% file: method.tex
\begin{figure*}[!t]
    \centering
    \includegraphics[width=\linewidth]{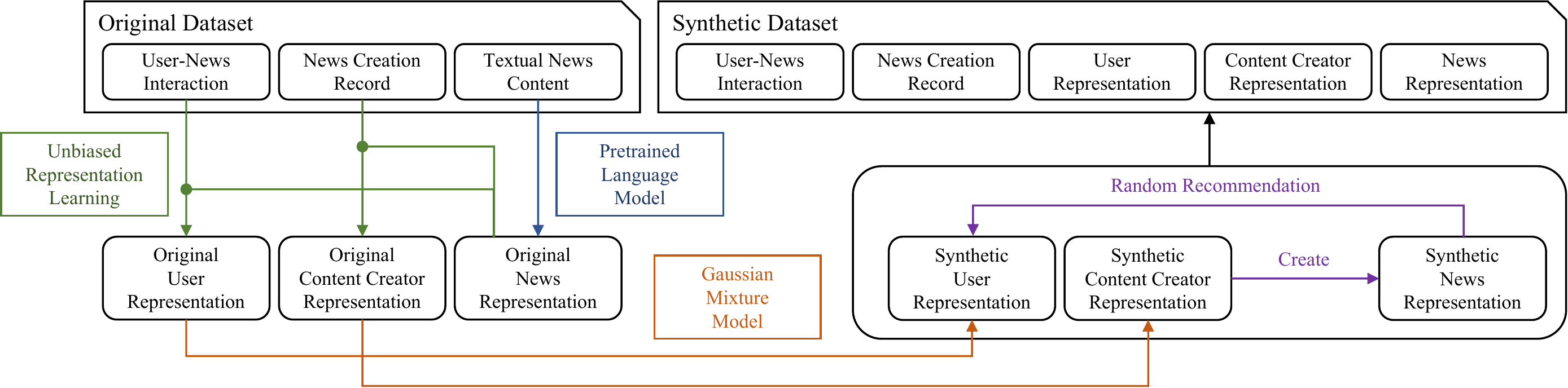}
    \caption{The procedure of synthetic data generation. This module transforms the original dataset into a synthetic dataset, where it resolves the inherent exposure bias and generates the latent space representations of users, content creators, and news.}
    \label{fig:generate}
\end{figure*}

\section{The Simulation Platform}
\label{sec:method}

This section introduces \ours, a simulation platform to dissect the evolution of news recommendation ecosystem. 
\ours consists of two key modules: 
\begin{tightitemize}
    \item {\bf 1. Synthetic data generation module}, in which \ours builds the latent space, deals with the inherent exposure bias of the original dataset and initializes the simulation;
    \item {\bf 2. Agent-based simulation module}, in which \ours builds the agents of users, content creators and recommender system, and simulates the recommendation feedback loops. 
\end{tightitemize}

\subsection{Synthetic Data Generation}
\label{sec:method-subsec:generation}

As the public datasets cannot meet all the specific requirements, especially the ground-truth user preference and the corresponding news features which are essential in simulating user decision process, we first generate synthetic data to initialize the simulation.
More specifically, we learn a latent space based on the textual news contents and original interactions from the dataset. Then, we sample from this latent space to generate synthetic users and their interactions. As such, we can simulate the evolution from scratch and control the scale of the simulating platform by hyper-parameters. In addition, since all users evolved in the simulation are synthetic, there will not be privacy concerns in our platform. 
Figure~\ref{fig:generate} illustrates the procedure of synthetic data generation.

\subsubsection{Building the Latent Space}
\label{sec:method-subsec:generation-subsubsec:plm}

The latent space is designed to model the user decision process. 
Each vector in the latent space represents the preference of one user or the features of one news. 
Here, building a latent space which could accurately reflect the human behaviors is the primary challenge. 
Recent success of the pretrained language models (PLMs)~\cite{pennington2014glove, devlin2018bert} provides us with a new solution for tackling this challenge. 
Pretrained on large-scale unlabeled corpus via self-supervision, the PLMs can learn to encode universal textual information, preserving key properties reflecting human cognition, e.g., word analogies could be solved with the vector arithmetic on top of word encodings~\cite{allen2019analogies, ethayarajh2019towards}. 
Thus we propose to take the semantic space of PLMs as the latent space. 
Using PLMs, news could be encoded directly in the latent space. 
For the users and content creators, we map them into the latent space via representation learning and vector arithmetic using the historical interactions. 
Thereby, we can provide textual explanation for every latent vector by retrieving similar vectors, making the evolution more explainable.

\subsubsection{Learning Unbiased User and Content Creator Latent Representation}
\label{sec:method-subsec:generation-subsubsec:unbiased}

The historical interactions of the users are derived from online algorithmic recommendations, i.e., user behaviors have been affected by the original recommender system of the news platform, resulting in the problem of exposure bias~\cite{chen2020bias, huang2020keeping}. 
We leverage the Inverse Propensity Score (IPS)~\cite{imbens2015causal} to deal with the exposure bias.
Formally, denoting the historical interactions between the users and the news as $I^U$ and the PLM-encoded news latent representation matrix as $H^N$, the goal is to learn the latent matrix $H^U$ of the users via unbiased bayesian personalized ranking~\cite{saito2020unbiased} using $I^U$ and $H^N$. The optimization objective is formulated as follows: 
\begin{equation}
    \label{equation:unbiased-learning-obj}
    \max_{H^U} \frac{1}{\left|\mathcal{D}_{\text {pair}}\right|} \sum_{(u, i, j) \in \mathcal{D}_{\text{pair}}} \omega_{u,i,j} \cdot ln \sigma \left(S_{u, i} - S_{u, j} \right) - \lambda\|H^U\|_2^2, 
\end{equation}
where 
$\mathcal{D}_{\text {pair}}$ refers to the pair-wise dataset extracted from the historical interactions via negative sampling. 
$S_{u, i} = H^U_u \cdot H^N_i$ is the matching score based on the latent representations. 
$\lambda$ is the regularization parameter. 
$\omega_{u,i,j} = \frac{I^U_{u, i}}{\theta_{u, i}}\left(1-\frac{I^U_{u, j}}{\theta_{u, j}}\right)$ , where $\theta_{u, i}$ is the inverse propensity score debiasing the exposure probability, which is defined as follows:
\begin{equation}
    \label{equation:unbiased-learning-ips}
    \theta_{u, i}=P\left(I^U_{u, i}=1 \mid \sigma \left(S_{u, i}\right)=1 \right).
\end{equation}
As news creation is not affected by the recommendation algorithms directly, we take the average of news latent vectors to form the latent representation matrix $H^C$ of content creators.

\subsubsection{Generating Synthetic Users and Content Creators}
\label{sec:method-subsec:generation-subsubsec:gmm&init}

Since the online news platforms usually come with multiple news panels, the original latent representations exhibit the distributional characteristic of a superposition of multiple Gaussian distributions. 
Correspondingly, we utilize the Gaussian Mixture Model (GMM)~\cite{bishop2006pattern} to fit the distributions and generate synthetic users. 
The optimization objective is formulated as: 
\begin{equation}
    \label{equation:gmm-objective}
    \max_{\gamma, \theta }   \sum_{i=1}^{N}log \left( \sum_{t=1}^{T} \gamma_i \phi ( H^U_i | \mathcal{N}(\mathbf{\mu}_t, \mathbf{\Sigma}_t ) ) \right).
\end{equation}
$\phi ( \cdot | \mathcal{N}(\mathbf{\mu}_t, \mathbf{\Sigma}_t ) )$ is the density function of the $t$-th multivariate Gaussian distribution. 
$N$ is the configurable numbers of users, which controls the scale of simulation.
$T$ is the number of categories in the original dataset. 
We take two steps to generate synthetic user $i$:
\begin{itemize}
\item {Step 1}: sample the category $\hat{t} \sim Multinomial(\gamma)$.
\item {Step 2}: sample the latent vector $\hat{H}^U_i \sim \mathcal{N}(\mathbf{\mu}_{\hat{t}}, \mathbf{\Sigma}_{\hat{t}} )$. 
\end{itemize}
Then, $\hat{H}^U_i$ is used as the real preference of synthetic user $i$ in the simulation. The generation of synthetic content creators follows the same procedure, so that the details are omitted.

\subsubsection{Initialization for Simulation}

To initialize the simulation, each synthetic content creator first samples $K$ news from the distribution $\mathcal{N}(\hat{H}^C_i, \rho^C_i \cdot \delta)$ to build the news pool. 
$\rho^C \in \mathbb{R}^M$ is a vector with elements ranging from $0$ to $1$, representing the degree of concentration of the content creators. 
$M$ is the configurable number of content creators. 
$\delta$ is a hyper-parameter which controls the magnitude of the latent space variation. 
Then the recommender system generates random recommendation lists $\mathcal{L}$, which is the default strategy when lacking historical interactions as training data. Note that the random recommendations here can also avoid the exposure bias issue during the initialization of the simulation.
In the end, the synthetic users respond to the recommendations according to the action policies defined in Section~\ref{sec:method-subsec:simulation-subsubsec:user}.

\begin{figure}[!t]
    \centering
    \includegraphics[width=\linewidth]{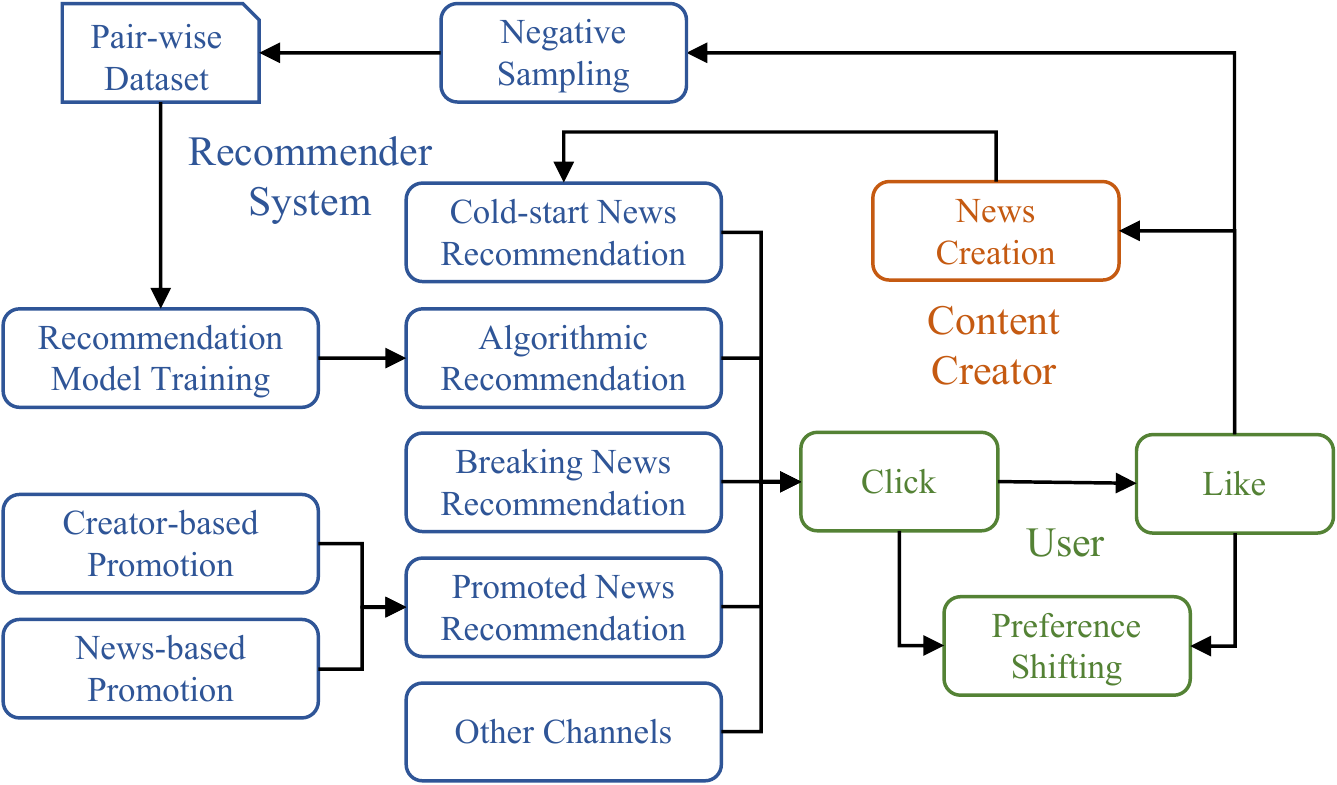}
    \caption{The procedure of the simulation framework. The agents of content creators, the recommender system, and users take actions in turn, constituting the simulation of recommendation feedback loops.}
    \label{fig:simulation}
\end{figure}

\subsection{Agent-based Simulation}
\label{sec:method-subsec:simulation}

In order to model the evolution process of NRE as in detail as possible, we attempt to build a comprehensive and extensible simulation system as illustrated in Figure~\ref{fig:simulation}. Next, we describe the detailed designs for the agents.

\subsubsection{Agent of Content Creator}
\label{sec:method-subsec:simulation-subsubsec:creator}

When creating news, the content creators leverage the greedy mechanism, which means that topics received more \texttt{LIKE}s in the last round would be more likely selected as the cornerstone of the future news. 
Formally, the probability of content creator $i$ generating news $j$ is defined as:
\begin{equation}
    \label{equation:creator-pickprob}
    P^C_{i, j} = \underset{j \in N^C_i}{\text{softmax}}\ \exp\{-\frac{1-\rho^C_i}{D^N_j+\varepsilon}\}, 
\end{equation}
where 
$N^C_i$ refers to the news created by content creator $i$. 
$\rho^C$ is the concentration vector of the content creators, controlling their sensitivity to the number of received \texttt{LIKE}s. 
$D^N_j$ is the number of received \texttt{LIKE}s of news $j$. 
$\varepsilon$ is a small constant to avoid zero in the denominator. 
Through sampling with replacement, each content creator selects $K$ news. 
Denoting the latent vector of the $k$-th news selected by content creator $i$ as $\hat{H}^{N_{\text{selected}}}_{\ i,k }$, 
the news is sampled from $\mathcal{N} \left( \hat{H}^{N_{\text{selected}}}_{\ i,k }, \rho^C_i \cdot \delta \right)$.
The number of \texttt{LIKE}s is diminishing marginally correlated with the income, and the creators with higher income would create news with higher quality due to more sufficient budget. 
Thus we define the quality of news as $log(D^C_i+1)$, 
where $D^C_i$ is the number of \texttt{LIKE}s received by creator $i$ in the last round (randomly drawn from Binomial distribution $\mathcal{B}(n_{\text{click}}, p_{\text{like}})$ for the initial round, where $n_{\text{click}}$ and $p_{\text{like}}$ are hyper-parameters). 
The created news will be passed to the recommender system and be recommended using cold-start recommendation strategies. Note that each news will be active for only $S$ rounds, reflecting the timeliness of news recommendation.

\subsubsection{Agent of Recommender System}
\label{sec:method-subsec:simulation-subsubsec:recsys}

The algorithmic recommendation and cold-start recommendation are the two basic elements of the recommender system agent.
To provide personalized algorithmic recommendations, the recommender system agent first learns the user preference in the embedding space using recommendation algorithms, e.g., BPR~\cite{rendle2012bpr}, from historical interaction datasets. 
However, due to the uncertainty of user actions and the time limit for news activation, the dataset is not guaranteed to cover all the users. 
Hence, for those covered by the dataset, we recommend the top-scored news using the trained recommendation model. 
For users not in the dataset, we recommend randomly. 
The newly created news could not involve in the algorithmic recommendation due to lacking interaction records. 
The agent applies strategies such as random recommendations and news from historically liked creators, to recommend cold-start news. 
\ours also supports extensive news recommendation strategies, such as breaking news, content creator-based promotion, and topic-based promotion, via independent exposure quotas. 
In the end, the agent merges the recommended news from all channels to form the final recommendation list $\mathcal{L}$.

\subsubsection{Agent of User}
\label{sec:method-subsec:simulation-subsubsec:user}

In the real-world news consumption scenarios, 
the users first decide whether to click on the news based on whether the title and abstract can draw their interests. After reading the news, the users will decide whether to give a \texttt{LIKE} according to if the content and quality of the news exceed their expectations~\cite{ma2018entire}.
We formulate the \texttt{CLICK} action as a probabilistic selection process~\cite{wilson2014humans, gershman2018deconstructing}, and the probability of user $i$ clicking on news $j$ is defined as: 
\begin{equation}
    \label{equation:user-clickprob}
    P^U_{i, j} = \underset{j \in \mathcal{L}_i}{\text{softmax}}\ \left( \rho^U_i \cdot \frac{\hat{H}^U_i \cdot \hat{H}^N_j}{\|\hat{H}^U_i\|_2^2 \cdot \|\hat{H}^N_j\|_2^2} \right), 
\end{equation}
where $\rho^U$ is the concentration vector of the users, controlling users' willingness to explore new content.
We define the degree to which user $i$ would like news $j$ as the reading utility $\mathcal{U}_{i, j}$, which is determined by two factors: the content utility and the quality utility. User $i$ will like news $j$ if the reading utility $\mathcal{U}_{i, j}$ exceeds user $i$'s expected threshold $\mathcal{T}_i$, which is formulated as:
\begin{equation}
    \label{equation:user-likecond}
    \mathcal{U}_{i, j} = \alpha_i \frac{\hat{H}^U_i \cdot \hat{H}^N_j}{\|\hat{H}^U_i\|_2^2 \cdot \|\hat{H}^N_j\|_2^2} + (1-\alpha_i) \frac{Q_j}{\mathcal{Q}} > \mathcal{T}_i,
\end{equation}
where $\alpha_i$ is a hyper-parameter to balance user $i$'s preference between content utility and quality utility. 
$Q_j$ refers to the quality of news $j$. 
$\mathcal{Q} = log(K \cdot N+1)$ is the highest quality the creators could obtain under certain simulation environment.
Based on the positive or negative interactions, the users strengthen or weaken their preference on the topics, respectively. 
We leverage the user-drift model~\cite{bountouridis2019siren} to formulate this process as follows: 
\begin{equation}
    \label{equation:user-shifting}
    \hat{H}^U_i \gets \hat{H}^U_i + \underset{j \in \mathcal{L'}_i }{\sum}  \mathbb{I}_{i, j} \cdot \frac{\delta(\hat{H}^N_j - \hat{H}^U_i)}{\|\hat{H}^N_j - \hat{H}^U_i\|_2^2} \cdot \exp\{-\frac{\rho^U_i}{\left \lfloor \mathcal{U}_{i, j} \right \rfloor_{0} + \varepsilon}\},    
\end{equation}
where $\mathcal{L'}_i$ is the set of news clicked by user $i$. 
$\mathbb{I}_{i, j}$ is $1$ if user $i$ likes news $j$ and $-1$ otherwise. 
$\left \lfloor \cdot \right \rfloor_{0}$ refers to the clipping operation with the minimal value of $0$.

%% file: experiments.tex
\section{Experiments}
\label{sec:experiments}

This section first describes the implementation details of our simulation experiments.
Then we introduce the analysis framework and present the simulation results. 
We attempt to answer the three research questions via life-cycle analysis, key factors and affecting mechanisms analysis, and recommender system designing strategies experiments:
\begin{tightitemize}
    \item What are the characteristics of each phase of the evolutionary lifecycle of NREs? (Findings 1\textasciitilde 6)
    \item What are the factors driving the evolution of NREs and how do these factors affect the evolutionary process? (Findings 7)
    \item How can we achieve better long-term multi-stakeholder utilities and avoid communities from falling into decline through the design strategy of recommender systems? (Findings 8)
\end{tightitemize}

\subsection{Implementation Details}
\label{sec:experiments-subsec:setting}

\subsubsection{Dataset}
\label{sec:experiments-subsec:setting-subsub:dataset}
We adopt Adressa~\cite{gulla2017adressa}, a Norwegian real-world news recommendation dataset covering one week of the whole web traffic (including 1,060,341 clicking behaviors from 133,765 users, and 14,661 news from 2,611 authors) from February 2017 on the \textit{www.adressa.no} website, as the original dataset for the consideration of data completeness~\footnote{Other high-quality news recommendation datasets, such as MIND, fail to meet our requirements due to the lack of author information.}. 

\subsubsection{Synthetic Data Generation}
\label{sec:experiments-subsec:setting-subsub:genetation}
We utilize BPEmb~\cite{heinzerling2018bpemb}~\footnote{We choose the off-the-shelf technique for engineering convenience, as the designing choices of different PLMs or collaborative filtering-based recommendation algorithms are orthogonal to our main contributions and would not affect the main findings.\label{choice}}, which is a collection of pre-trained multilingual subword embeddings based on Byte-Pair Encoding (BPE) and trained on Wikipedia, to encode the original news. 
The number of synthetic users and content creators is $10000$ and $1000$, respectively. 
The initial user \texttt{LIKE} probability $p_{\text{like}}$ is set as $0.1$.
We simulate $100$ rounds and in each round every creator produces $5$ news. 

\subsubsection{Agent-based Modeling}
\label{sec:experiments-subsec:setting-subsub:modeling}
The default recommender system performs algorithmic recommendations using the BPR algorithm~\cite{rendle2012bpr}~\textsuperscript{\ref{choice}} and uses random recommendation for cold-start scenarios. 
The algorithmic model is trained using the Adam optimizer on the pair-wise dataset. 
We take \texttt{LIKE}s as the positive records and sample negative records uniformly with a negative sampling ratio of $1:9$~\footnote{Although sampling negative records from clicked news or exposed news could yield better algorithmic recommendation coverage on the news, the improved coverage comes from incorporating more news as negative samples into the process of model training, of which the learned embeddings present no practical values.}. 
The learning rate is $1e-3$ and the batch size is $1024$. 
The model would be trained for up to $100$ epochs with an early stop strategy where the threshold number of non-exceeding step is set as $3$. 
The dataset is split as subsets for training, evaluation and testing randomly with the ratio of $80\%$, $10\%$ and $10\%$. 
The evaluating metric is $MRR@5$. 
Each user is recommended with $100$ news ($80$ news from algorithmic recommendation, $20$ news from cold-start recommendation) in each round and the number of clicks $n_{\text{click}}$ is set as $10$. 
The user thresholds $\mathcal{T}$ are sampled from $\mathcal{N} \left( 0.3, 0.1 \right)$. 
The user utility preference $\alpha$ is sampled from $\mathcal{N} \left( 0.5, 0.1 \right)$. 
The user and creator concentration $\rho^U$ and $\rho^C$ are sampled from $\mathcal{N} \left( 0.5, 0.1 \right)$. 
The above four hyper-parameters are clipped to be non-negative.

\subsection{Analysis Framework and Simulation Results}
\label{sec:experiments-subsec:process}

We analyze the simulation results from three perspectives: quantitative metrics, latent space visualization and latent space explanation.

\begin{figure*}[!t]
    \centering
    \includegraphics[width=0.75\linewidth]{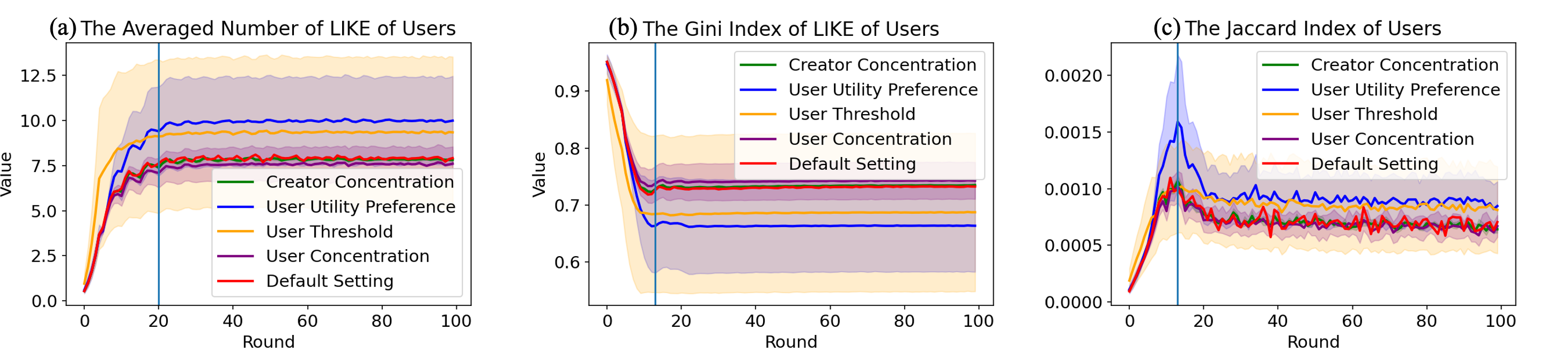}
    \caption{User related metrics vs. rounds, including average number of \texttt{LIKE}s, Gini index of users and  Jaccard index of users.}
    \label{fig:metrics-track-user}
\end{figure*}

\begin{figure*}[!t]
    \centering
    \includegraphics[width=\linewidth]{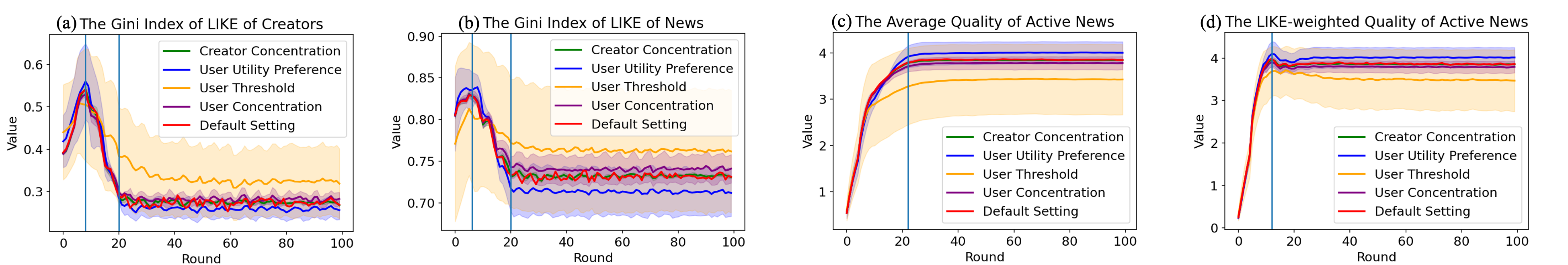}
    \caption{Content creator related metrics vs. rounds, including the Gini index of creators and news, the average quality of active news and the \texttt{LIKE}-weighted quality of active news.}
    \label{fig:metrics-track-creator}
\end{figure*}

\begin{figure*}[!t]
    \centering
    \includegraphics[width=\linewidth]{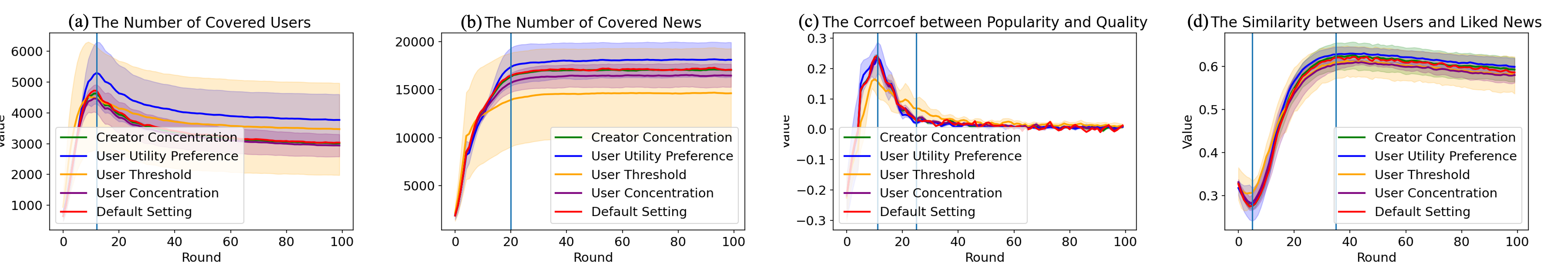}
    \caption{Recommender system related metrics vs. rounds, including the number of users and news covered by algorithmic recommendation,  correlation coefficient between popularity and quality, and cosine similarity between users and liked news.}
    \label{fig:metrics-track-recsys}
\end{figure*}

\subsubsection{Quantitative Metrics}
\label{sec:experiments-subsec:process-subsubsec:metrics}

To comprehensively understand the evolutionary process, we summarize and extend the quantitative metrics used in the existing literature from the following five perspectives: 
{\bf 1) Interaction}, including the number of \texttt{LIKE}s and their Gini index~\cite{chaney2018algorithmic, sun2019debiasing}. A lower Gini index represents better fairness; 
{\bf 2) Coverage}, including the numbers of users and news covered by the algorithmic recommendation; 
{\bf 3) Quality}, including average quality of active news, \texttt{LIKE}-weighted average quality of active news and the Pearson product-moment correlation coefficient between news quality and the number of \texttt{LIKE}s; 
{\bf 4) Homogenization}, including the Jaccard index~\cite{chaney2018algorithmic, aridor2020deconstructing}, of which a higher value represents a higher degree of overlap in news reading among the users; 
{\bf 5) Latent representation}, including the cosine similarity between the latent representations of users and their liked news. 

Figure~\ref{fig:metrics-track-user},~\ref{fig:metrics-track-creator},~\ref{fig:metrics-track-recsys} present the evolution of these metrics under varying environmental hyper-parameters, including $\mathcal{T}$ (user threshold), $\alpha$ (user utility preference), $\rho^U$ (user concentration) and $\rho^C$ (content creator concentration). The lines indicate the mean values, and the shaded areas represent the variance. 
From Figure~\ref{fig:metrics-track-user},~\ref{fig:metrics-track-creator},~\ref{fig:metrics-track-recsys} we can observe that with the 10th and 20th rounds as the approximate boundaries, all the metrics under any environmental hyper-parameter present a two-stage or three-stage pattern. 


\textbf{Findings 1:} Recommender system-driven online news communities naturally exhibit the ``start-up''-``growth''-``maturity\&decline'' life-cycle under a variety of user group settings.

\begin{figure}[!t]
    \centering
    \includegraphics[width=\linewidth]{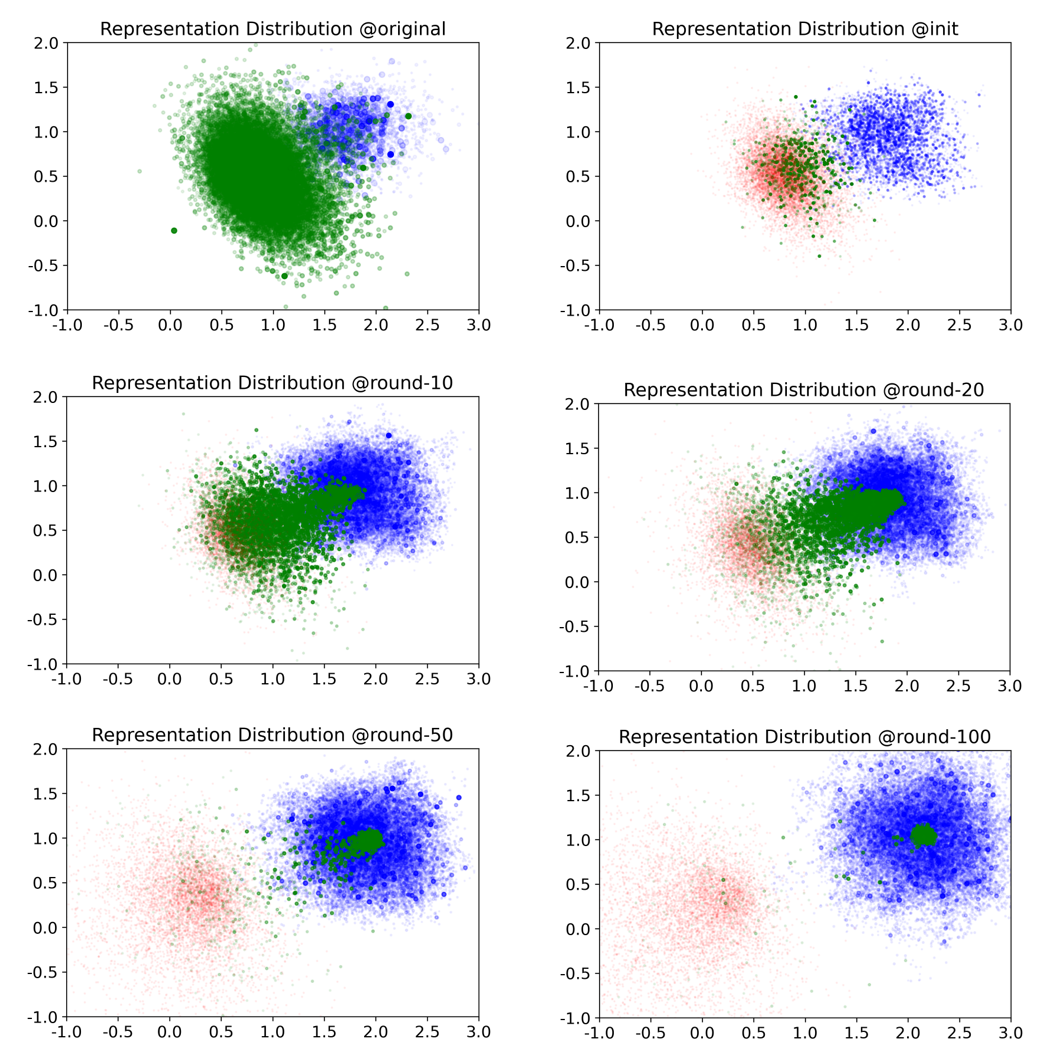}
    \vspace{-8pt}
    \caption{The evolution of users and news in the latent representation space. The news is colored with blue, users with \texttt{LIKE} records are colored with green, and users without \texttt{LIKE} records are colored with red. }
    \label{fig:latent-distribution}
    \vspace{-5pt}
\end{figure}

\subsubsection{Latent Space Visualization}
\label{sec:experiments-subsec:process-subsubsec:visualization}

Figure~\ref{fig:latent-distribution} visualizes the evolution of the latent representations of users and news in the default simulation setting using PCA~\cite{bishop2006pattern}. The news is colored with blue, users with \texttt{LIKE} records are colored with green and users without \texttt{LIKE} records are colored with red. The node size represents the number of \texttt{LIKE}s. 
Although the quantitative metrics exhibit a multi-stage pattern, the evolutionary trend of the latent space representations is consistent, i.e., users gradually differentiate into in-the-loop and out-the-loop. Users in the loop form a stable community with convergent interests, while users out the loop present fragmented interests. 
During the interaction between the 10th and 20th rounds, the users have basically completed the differentiation, which indicates that the growth phase is critical to user engagement. 


\textbf{Findings 2:} Recommender system-driven online news communities would inevitably yield a convergence of community topics and lead to user differentiation. The key period that determines user engagement is the growth phase.

\begin{table*}[!t]
\label{table:case}
\caption{The textual explanation of the latent representation evolution of six synthetic users, where the first three users participated in the recommendation feedback loops, and the last three users failed to get involved. In-the-loop users gradually develop a convergent interest in big topics, while out-the-loop users retain their personalized interests in minority topics.}
\centering
\begin{tabular}{|ll|llllll|}
\hline
\multicolumn{2}{|l|}{user}                                                  & initialization & 5-th round  & 10-th round & 20-th round & 50-th round & 100-th round \\ \hline
\multicolumn{1}{|l|}{\multirow{9}{*}{in the loop}}  & \multirow{3}{*}{user \#1} & November       & Autumn      & Year        & Job         & Job         & Job          \\
\multicolumn{1}{|l|}{}                          &                           & February       & Manager     & Example     & Year        & Year        & Year         \\
\multicolumn{1}{|l|}{}                          &                           & Actors         & Child       & Autumn      & Europe      & Europe      & Europe       \\ \cline{2-8} 
\multicolumn{1}{|l|}{}                          & \multirow{3}{*}{user \#2} & Oslo           & Oslo        & Mountain    & Mountains   & Europe      & Europe       \\
\multicolumn{1}{|l|}{}                          &                           & School         & School      & Norway      & Wood        & Time        & Time         \\
\multicolumn{1}{|l|}{}                          &                           & Post Office    & Post Office & Stations    & Norway      & Reason      & Reason       \\ \cline{2-8} 
\multicolumn{1}{|l|}{}                          & \multirow{3}{*}{user \#3} & Nuts           & Nuts        & Sky         & Time        & Year        & Year         \\
\multicolumn{1}{|l|}{}                          &                           & Christian      & Christian   & Knowledge   & Reason      & Reason      & Reason       \\
\multicolumn{1}{|l|}{}                          &                           & Cello          & Cello       & Musician    & Example     & Example     & Example      \\ \hline
\multicolumn{1}{|l|}{\multirow{9}{*}{out the loop}} & \multirow{3}{*}{user \#4} & Hometown       & Hometown    & Athletes    & Athletes    & Athletes    & Athletes     \\
\multicolumn{1}{|l|}{}                          &                           & Athletes       & Athletes    & Hometown    & Hometown    & Ferret      & Ferret       \\
\multicolumn{1}{|l|}{}                          &                           & Finals         & Prophets    & Ferret      & Ferret      & Ropes       & Knife        \\ \cline{2-8} 
\multicolumn{1}{|l|}{}                          & \multirow{3}{*}{user \#5} & Runners        & Runners     & Runners     & Tea         & Tea         & Tea          \\
\multicolumn{1}{|l|}{}                          &                           & Champions      & Champions   & Tea         & Runner      & Runner      & Vascular     \\
\multicolumn{1}{|l|}{}                          &                           & Tea            & Tea         & Karate      & Karate      & Vascular    & Frozen       \\ \cline{2-8} 
\multicolumn{1}{|l|}{}                          & \multirow{3}{*}{user \#6} & Money          & Countryman  & Billing     & Billing     & Billing     & Billing      \\
\multicolumn{1}{|l|}{}                          &                           & Countryman     & Money       & Countryman  & Countryman  & Countryman  & Countryman   \\
\multicolumn{1}{|l|}{}                          &                           & Bill           & Bill        & Money       & Police      & Police      & Police       \\ \hline
\end{tabular}
\end{table*}

\subsubsection{Textual Latent Space Explanation}
\label{sec:experiments-subsec:process-subsubsec:explanation}

Since we construct the latent space through PLMs, every vector in the space could be interpreted textually using words with similar embeddings. This helps to understand the evolution of the individual users through case studies, as a complement to the latent space visualization. 
We randomly select $3$ users separately from in and out the recommendation feedback loop, and Table~$1$ presents the detailed textual explanation of the evolution of the $6$ users~\footnote{For each user, we retrieve the top $3$ nouns with the highest cosine similarity 
from the wordbase which contains $200,000$ Norwegian sub-words, and translate them into English as the textual explanation of the target representation.}. 

For users in the loop, the most similar words evolve toward more abstract, broader, and generalized nouns, e.g., from ``Actors'' to ``Job'' and from ``Oslo'' to ``Norway'' and ``Europe''. The evolutionary speed of different users varies, but all converge by the $50$-th round. This phenomenon reflects the migration of user preference from the personalized niche topics to the trending topics that are widely discussed on the platform, as a consequence of continuously interacting with the recommender system. 
For users out the loop, their representations shift slightly, but always focus on the specific and personalized topics. For instance, users \#4\textasciitilde\#6 keep their interests towards ``Athletes'', ``Tea'' and ``Bill'' respectively throughout the simulation. 


\textbf{Findings 3:} In recommender system-driven online news communities, users' personalized interests are assimilated in the process of continuous interaction with the recommendation system.

\subsection{Life-Cycle Analysis}
\label{sec:experiments-subsec:lifecycle}

To the best of our knowledge, \ours is the first work to validate the lifecycle of NREs through simulation. In this section, we discuss the characteristics of each phase in detail.

\subsubsection{Start-up Phase}
\label{sec:experiments-subsec:lifecycle-subsubsec:startup}

Corresponding to the first 10 rounds, this phase addresses the cold-start user issue, i.e., 
involving the users in the recommendation feedback loop via random recommendations. 
The \texttt{LIKE}s in this stage are mainly driven by the news quality, leading to the positive correlation between quality and popularity as shown in Figure~\ref{fig:metrics-track-recsys} (c). 
The development of this phase is catalyzed by two main factors: 
1) the quality feedback loop, which refers to the mutual promotion between quality and popularity based on the positive correlation; and 
2) the interest-quality confusion, which refers to that until enough data is accumulated to accurately estimate user interests, the recommendation algorithm confuses the quality-driven \texttt{LIKE}s as interest-driven. 
This helps the popular content creators to obtain excessive exposure and further enhances the quality feedback loop, resulting in the decline in the similarity between the latent vectors of users and their liked news as shown in Figure~\ref{fig:metrics-track-recsys} (d). 
Most users can benefit from the improved news quality, decreasing the Gini index of user \texttt{LIKE}s as shown in Figure~\ref{fig:metrics-track-user} (b). 


\textbf{Findings 4:} In start-up phase, NREs accumulate data for estimating user interests from random recommendations and high-quality news, which can address the cold-start user issue. The quality feedback loop and interest-quality confusion contribute to the emergence of highly popular content creators via excessive exposure.

\subsubsection{Growth Phase}
\label{sec:experiments-subsec:lifecycle-subsubsec:growth}
With the accumulation of data, the recommendation model becomes increasingly accurate in estimating user interests, and the \texttt{LIKE} behaviors are more interest-driven than quality-driven. The correlation between quality and popularity gradually weakens, resulting in the end of quality feedback loop around the 10th round. 
As shown in Figure~\ref{fig:latent-distribution} (c), the density of news around each in-the-loop user is uneven, with a higher density in the direction towards the mainstream news topic and smaller densities in other directions. As a result, the liked news is statistically closer to the mainstream news topic compared with the users in the latent space, and the user interests are gradually cultivated towards the mainstream news topic through continuous interactions with the recommender system. 
On the contrary, the out-the-loop users are trapped in the deadlock of ``no \texttt{LIKE} - not covered by the algorithm - poor recommendation quality - no \texttt{LIKE}''. They still give \texttt{LIKE}s occasionally, but the time-bound data is too sparse for the recommendation models to accurately estimate their interests and stimulate more \texttt{LIKE}s in future rounds.
More frequent and balanced \texttt{LIKE}s stimulate the growth of news quality, but due to the declining popularity of high quality news, the quality weighted by the number of \texttt{LIKE}s remains stable as shown in Figure~\ref{fig:metrics-track-creator} (b) (c) (d). 
As the quality feedback loop terminates, content creators can't receive excessive attention and thus the news quality falls back, resulting in that quality-sensitive users stop giving \texttt{LIKE}s which presents as the drop in user coverage as shown in Figure~\ref{fig:metrics-track-recsys} (a). 


\textbf{Findings 5:} In the growth phase, in-the-loop users evolve towards common topics under the effect of distribution bias, while out-the-loop users are trapped in a deadlock, resulting in user differentiation. The growing accurate algorithmic recommendation leads to the ending of the quality feedback loop, resulting in the reduced coverage of quality-sensitive users.

\subsubsection{Maturity\&Decline Phase}
\label{sec:experiments-subsec:lifecycle-subsubsec:maturity}
At around the 20th round, the NRE enters the maturity\&decline phase as most of the key metrics are relatively stable. 
In this phase, the in-the-loop users are dynamically kept in the bubble of common topics. 
Although their interests may shift towards the edge of the bubble due to clicking on some diverse news, they would be soon back to the center due to the news articles around them are highly similar to the center of the bubble. 
As shown in Figure~\ref{fig:metrics-track-creator} (a) (b), the Gini index of \texttt{LIKE}s on news is high but that on content creators is low, indicating that the news even created by the same creator presents highly varying popularity. In addition to the greedy creating mechanism, the process of news creation is highly stochastic and presents a natural tendency of expansion, which is reflected in Figure~\ref{fig:latent-distribution} by the expansion of blue nodes and the Figure~\ref{fig:metrics-track-recsys} (d) by a small decrease in user-news similarity. 
The declining user-news similarity further leads to the quitting of interest-sensitive users as shown in Figure~\ref{fig:metrics-track-recsys} (a).


\textbf{Findings 6:} In the maturity\&decline phase, in-the-loop users share common topics, around which content creators publish diverse news. With the constraints of the recommender system, the NRE maintains stability of the community but along with slow loss of interest-sensitive users.

\begin{figure}[!t]
    \centering
    \includegraphics[width=\linewidth]{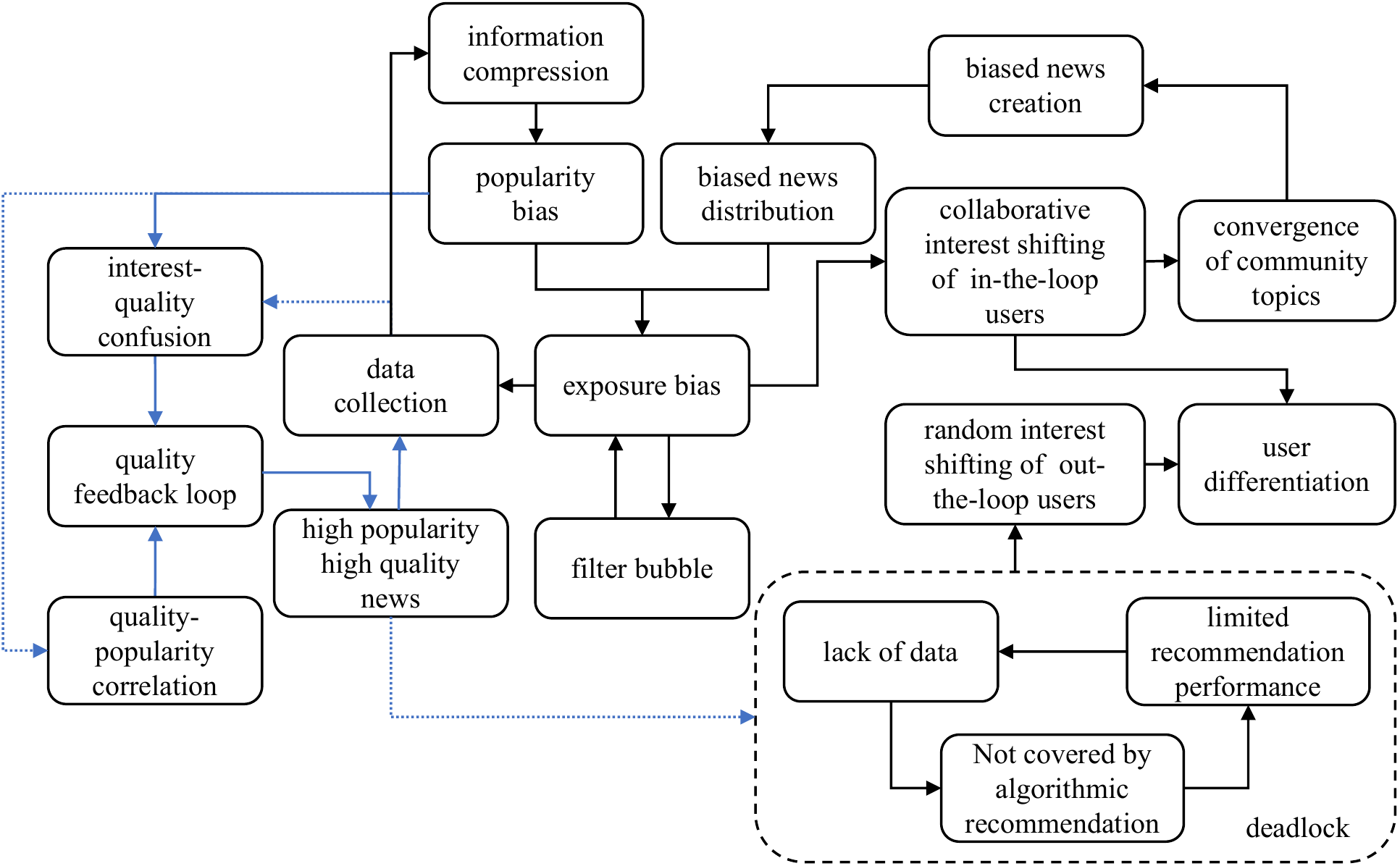}
    \caption{The relationship graph of news recommendation ecosystem evolution. The blue arrows occur in the start-up phase. The dotted arrows represent negative effects.}
    \label{fig:structure}
\end{figure}

\subsection{Key Factors and Affecting Mechanisms}
\label{sec:experiments-subsec:Mechanism}

Figure~\ref{fig:structure} illustrates the key factors and affecting mechanisms of NRE evolution, from which we could find that the re-emerging exposure bias and the deadlock are the direct causes of the different evolution trends between in-the-loop users and out-the-loop users, resulting in the user differentiation and convergence of topics.  

The re-emerging exposure bias is caused by multiple factors. 
1) According to the Information Theory~\cite{shannon1948mathematical}, recommendation algorithms can be interpreted as a process of information compression, inevitably leading to the popularity bias, in which high-frequency items are encoded and recommended more efficiently to improve system performance. Reflecting on the evolution of NREs, the widely discussed common topics would grab the exposure resources from personalized topics in the algorithmic recommendation. 
2) Due to the profit-seeking nature of content creators, they are more motivated to create news around topics of high public interest, which leads to a decreasing density of news distribution from common topics to personalized topics. Therefore, even when random recommendations are applied, NREs may evolve toward topic convergence due to the distribution bias.
3) Filter bubble and exposure bias promote each other and together lead to the subtle interest shift of users. Filter bubble refers to that algorithmic recommendations estimate user interests and recommend similar items based on historical user interactions. The restricted news exposure makes the exposure bias difficult to be perceived by the users. 

Besides, the popularity bias presents different effects in different evolutionary phases. In the start-up phase, with interest-quality confusion and quality-popularity correlation, the popularity bias is expressed as the enhanced exposure of high-quality news. With the accumulation of data and the rising performance of algorithmic recommendations, \texttt{LIKE} behaviors are growing interest-driven compared to quality-driven, debilitating interest-quality confusion and quality-popularity correlation. The popularity bias also gradually evolves from recommending high-quality news to simply recommending high-popularity news. During this process, some high-popularity high-quality news topics are cultivated, which play important roles in facilitating user engagement. 


\textbf{Findings 7:} Popularity bias, biased news distribution, and filter bubble together lead to exposure bias -- the key factor affecting user differentiation and topic convergence. The high-popularity\&quality news is crucial in breaking the deadlock of out-the-loop users.

\begin{figure*}[!t]
    \centering
    \includegraphics[width=0.75\linewidth]{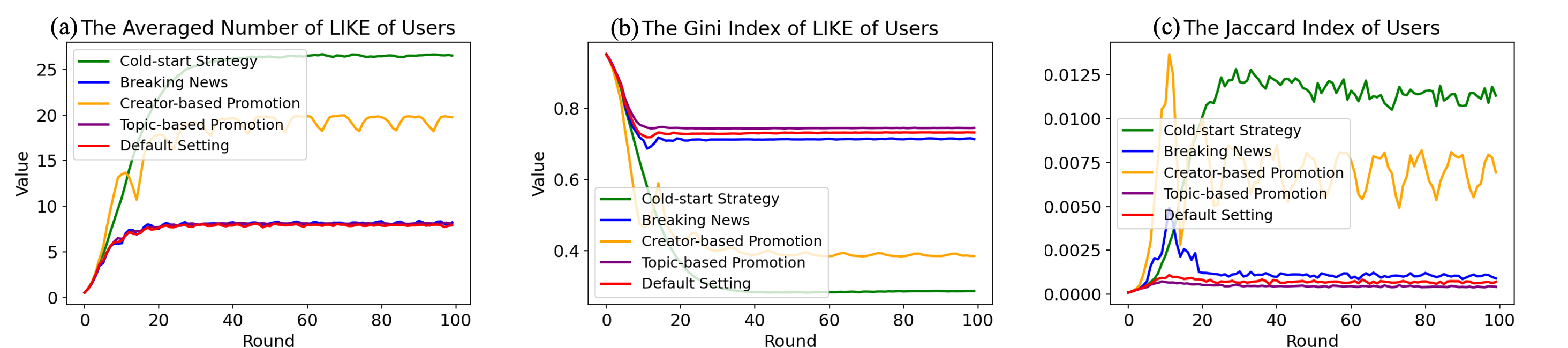}
    \caption{The user-related metrics under different recommender system designs, including the averaged number of \texttt{LIKE}s, the Gini index of users and the Jaccard index of users.}
    \label{fig:metrics-guide-user}
\end{figure*}

\begin{figure*}[!t]
    \centering
    \includegraphics[width=\linewidth]{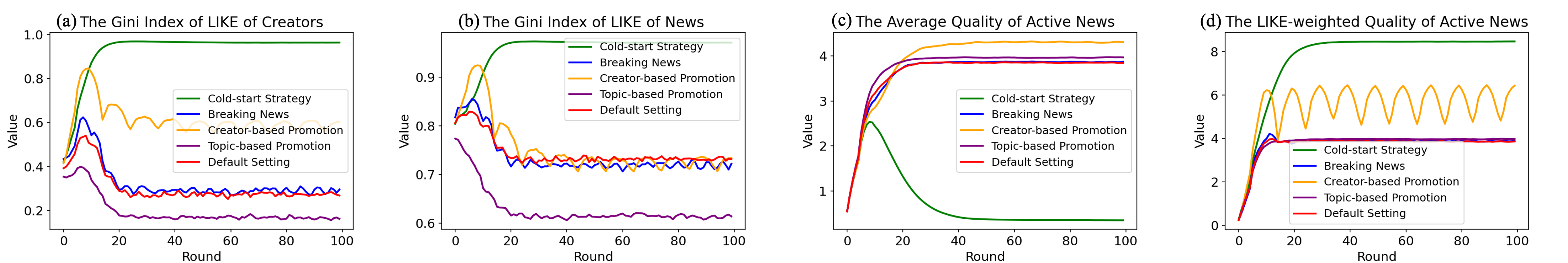}
    \caption{The content creator-related metrics under different recommender system designs, including the Gini index of content creators and news, the average quality of active news and the \texttt{LIKE}-weighted quality of active news.}
    \label{fig:metrics-guide-creator}
\end{figure*}

\begin{figure*}[!t]
    \centering
    \includegraphics[width=\linewidth]{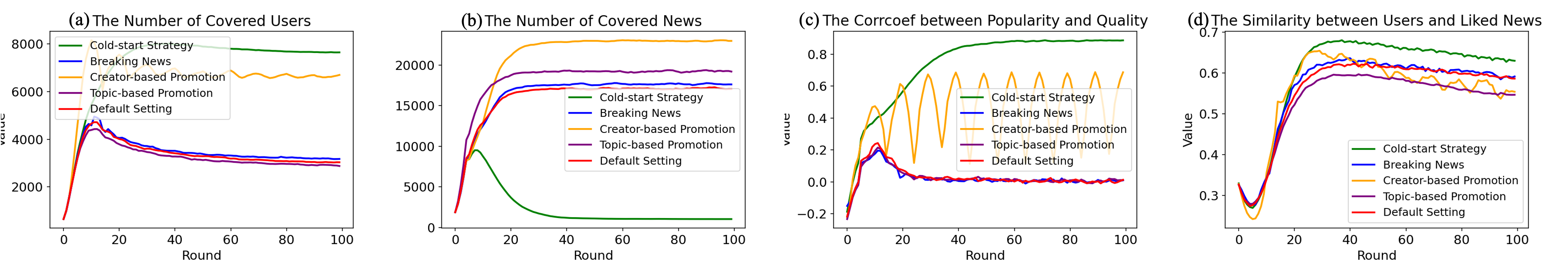}
    \caption{The recommender system-related metrics under different recommender system designs, including the number of users and news covered by the algorithmic recommendation, the correlation coefficient between popularity and quality, and the cosine similarity between users and their liked news.}
    \label{fig:metrics-guide-recsys}
\end{figure*}

\subsection{Improving the Evolution via Recommender System Designing Strategies}
\label{sec:experiments-subsec:improving}

This section explores how to utilize recommender system designing strategies to guide the evolution of NREs and thus achieve improved long-term utility as shown in Figure~\ref{fig:metrics-guide-user},~\ref{fig:metrics-guide-creator}, and~\ref{fig:metrics-guide-recsys}. 

\subsubsection{Cold-start}
\label{sec:experiments-subsec:improving-subsubsec:cold-start}

Illustrated by the green lines in Figure~\ref{fig:metrics-guide-user},~\ref{fig:metrics-guide-creator},~\ref{fig:metrics-guide-recsys}, we push the cold-start news to users who used to like the corresponding content creators instead of recommending randomly. 
This method attempts to form a stable cross-round exposure relationship between users and content creators, which enhances the quality feedback loop in the start-up phase.
However, this approach leads to a serious monopoly. The content creators who have not achieved a first-mover advantage are suppressed by the quality feedback loop, damaging the algorithmic coverage and average quality of news, which in turn makes the diversity of the ecosystem severely challenged.

\subsubsection{Breaking News}
\label{sec:experiments-subsec:improving-subsubsec:hot}

Illustrated by the blue lines in Figure~\ref{fig:metrics-guide-user},~\ref{fig:metrics-guide-creator},~\ref{fig:metrics-guide-recsys}, we add the $20$ most liked but unrecommended news in the previous round to the recommendation list, and correspondingly reduce the number of algorithmically recommended news by $20$. 
Relying on the positive correlation between popularity and quality, this method could provide users with news of higher quality, while ensuring user exploration and avoiding the monopoly due to the absence of a stable exposure relationship between users and content creators. 
From the perspective of exploitation and exploration, reading breaking news could also be regarded as a kind of user exploration, which can help mitigate the negative effects of the filter bubble. 
However, this method could not prevent the collapsing of the critical correlation between popularity and quality as discussed in Section~\ref{sec:experiments-subsec:Mechanism}, resulting in the declining effectiveness of recommending breaking news.

\subsubsection{Promotion}
\label{sec:experiments-subsec:improving-subsubsec:promotion}

Illustrated by the yellow and purple lines in Figure~\ref{fig:metrics-guide-user},~\ref{fig:metrics-guide-creator},~\ref{fig:metrics-guide-recsys}, we split $20$ quotas of news from algorithmic recommendations to the promotion of randomly selected content creators or topics. The promoted content creators or topics are reset per $10$ rounds. 
The content creator-based promotion builds a stable exposure relationship and leverages the quality feedback loop to cultivate high-popularity high-quality news. But different from the cold-start strategies, the promotion could be terminated proactively before the current quality feedback loop cultivates a harmful monopoly, thereby enhancing the user experience and creator creativity. As an independent news-feeding channel, it could mitigate the negative effects of the filter bubble. Besides, by rebuilding the quality feedback loop, it also directs the effect of popularity bias toward beneficial recommendations for high-quality news.

As we select the topics randomly when performing the topic-based promotion, the trending topics have an equal chance to be promoted as the personalized topics, so that the promotion has a relatively larger impact on the personalized topics considering their relatively lower exposures. Theoretically, it can be used to improve the engagements of out-the-loop users, but due to the quality of promoted news cannot always be promised, it is difficult to convert the exposures into \texttt{LIKE}s, resulting in limited effectiveness. 


\textbf{Findings 8:} Among the typical recommender system designing strategies, periodic content creator-based promotion is the most effective. By actively building the quality feedback loop, it could create waves of high-popularity high-quality news topics throughout the NREs, and meanwhile keep the monopoly under control via periodic reset.

\subsection{System Validation}
\label{sec:experiments-subsec:validation}

In this section, we compare the findings conducted from our simulating experiments with those of existing works, so as to illustrate the validity and advancement of \ours.

Findings 1\textasciitilde 3 verify the existence of a multi-phase life-cycle in the evolution of online news communities and highlight the phenomena of topic convergence, user differentiation, and interest assimilation during the evolution of communities, which is consistent with the findings of extensive empirical studies of online communities~\cite{iriberri2009life}.

Findings 4\textasciitilde 7 attempt to explain the evolutionary process of online news communities, which also validate and extend the conclusions of existing works.
Regarding the relationship between popularity and quality, Zhao \etal~\cite{zhao2021popularity} and Nematzadeh \etal~\cite{ciampaglia2018algorithmic} pointed out that under certain conditions, high quality leads to high popularity. In our Findings 4\textasciitilde 5, we discuss the quality feedback loops in the start-up and growth phases,  further revealing how this correlation establishes and expires. 
Liu \etal~\cite{liu2021interaction} and Bountouridis \etal~\cite{bountouridis2019siren} discussed the impact of recommendation algorithms on view convergence and user fragmentation, which are verified in our Findings 5. 
Yao \etal~\cite{yao2021measuring} found that the popularity bias of recommendation results is mitigated as more interactions occur, which is consistent with our Findings 6.
Chaney \etal~\cite{chaney2018algorithmic} proposed that recommender systems amplify the homogeneity of user content consumption because of algorithmic confounding, which is similar to our arguments about exposure bias in Findings 7.

Findings 8 explores how recommender system designs could contribute to the prosperity of online communities. Nematzadeh \etal~\cite{ciampaglia2018algorithmic} and Jiang \etal~\cite{jiang2019degenerate} highlighted the impact of user exploration on improving long-term utilities. We further refine platform-guided user exploration into strategies such as cold-start, breaking news, and promotions, which are tested and discussed in detail. The results show that a well-designed user exploration strategy can lead to stable improvement, but inappropriate explore strategies may play negative roles.

%% file: main.bbl
\begin{thebibliography}{10}
\providecommand{\url}[1]{#1}
\csname url@samestyle\endcsname
\providecommand{\newblock}{\relax}
\providecommand{\bibinfo}[2]{#2}
\providecommand{\BIBentrySTDinterwordspacing}{\spaceskip=0pt\relax}
\providecommand{\BIBentryALTinterwordstretchfactor}{4}
\providecommand{\BIBentryALTinterwordspacing}{\spaceskip=\fontdimen2\font plus
\BIBentryALTinterwordstretchfactor\fontdimen3\font minus
  \fontdimen4\font\relax}
\providecommand{\BIBforeignlanguage}[2]{{%
\expandafter\ifx\csname l@#1\endcsname\relax
\typeout{** WARNING: IEEEtran.bst: No hyphenation pattern has been}%
\typeout{** loaded for the language `#1'. Using the pattern for}%
\typeout{** the default language instead.}%
\else
\language=\csname l@#1\endcsname
\fi
#2}}
\providecommand{\BIBdecl}{\relax}
\BIBdecl

\bibitem{li2019survey}
M.~Li and L.~Wang, ``A survey on personalized news recommendation technology,''
  \emph{IEEE Access}, vol.~7, pp. 145\,861--145\,879, 2019.

\bibitem{wu2022personalized}
C.~Wu, F.~Wu, Y.~Huang, and X.~Xie, ``Personalized news recommendation: Methods
  and challenges,'' \emph{ACM Transactions on Information Systems (TOIS)},
  2022.

\bibitem{merchant2014click}
G.~Merchant, N.~Weibel, K.~Patrick, J.~H. Fowler, G.~J. Norman, A.~Gupta,
  C.~Servetas, K.~Calfas, K.~Raste, L.~Pina \emph{et~al.}, ``Click “like”
  to change your behavior: a mixed methods study of college students’
  exposure to and engagement with facebook content designed for weight loss,''
  \emph{Journal of medical Internet research}, vol.~16, no.~6, p. e3267, 2014.

\bibitem{baker2017rational}
C.~L. Baker, J.~Jara-Ettinger, R.~Saxe, and J.~B. Tenenbaum, ``Rational
  quantitative attribution of beliefs, desires and percepts in human
  mentalizing,'' \emph{Nature Human Behaviour}, vol.~1, no.~4, pp. 1--10, 2017.

\bibitem{liu2021interaction}
P.~Liu, K.~Shivaram, A.~Culotta, M.~A. Shapiro, and M.~Bilgic, ``The
  interaction between political typology and filter bubbles in news
  recommendation algorithms,'' in \emph{Proceedings of the Web Conference
  2021}, 2021, pp. 3791--3801.

\bibitem{mueller1972life}
D.~C. Mueller, ``A life cycle theory of the firm,'' \emph{The Journal of
  Industrial Economics}, pp. 199--219, 1972.

\bibitem{iriberri2009life}
A.~Iriberri and G.~Leroy, ``A life-cycle perspective on online community
  success,'' \emph{ACM Computing Surveys (CSUR)}, vol.~41, no.~2, pp. 1--29,
  2009.

\bibitem{stavinova2022synthetic}
E.~Stavinova, A.~Grigorievskiy, A.~Volodkevich, P.~Chunaev, K.~Bochenina, and
  D.~Bugaychenko, ``Synthetic data-based simulators for recommender systems: A
  survey,'' \emph{arXiv preprint arXiv:2206.11338}, 2022.

\bibitem{luo2022mindsim}
X.~Luo, Z.~Liu, S.~Xiao, X.~Xie, and D.~Li, ``Mindsim: User simulator for news
  recommenders,'' in \emph{Proceedings of the ACM Web Conference 2022}, 2022,
  pp. 2067--2077.

\bibitem{geschke2019triple}
D.~Geschke, J.~Lorenz, and P.~Holtz, ``The triple-filter bubble: Using
  agent-based modelling to test a meta-theoretical framework for the emergence
  of filter bubbles and echo chambers,'' \emph{British Journal of Social
  Psychology}, vol.~58, no.~1, pp. 129--149, 2019.

\bibitem{jiang2019degenerate}
R.~Jiang, S.~Chiappa, T.~Lattimore, A.~Gy{\"o}rgy, and P.~Kohli, ``Degenerate
  feedback loops in recommender systems,'' in \emph{Proceedings of the 2019
  AAAI/ACM Conference on AI, Ethics, and Society}, 2019, pp. 383--390.

\bibitem{aridor2020deconstructing}
G.~Aridor, D.~Goncalves, and S.~Sikdar, ``Deconstructing the filter bubble:
  User decision-making and recommender systems,'' in \emph{Fourteenth ACM
  Conference on Recommender Systems}, 2020, pp. 82--91.

\bibitem{sun2019debiasing}
W.~Sun, S.~Khenissi, O.~Nasraoui, and P.~Shafto, ``Debiasing the
  human-recommender system feedback loop in collaborative filtering,'' in
  \emph{Companion Proceedings of The 2019 World Wide Web Conference}, 2019, pp.
  645--651.

\bibitem{zhao2021popularity}
Z.~Zhao, J.~Chen, S.~Zhou, X.~He, X.~Cao, F.~Zhang, and W.~Wu, ``Popularity
  bias is not always evil: Disentangling benign and harmful bias for
  recommendation,'' \emph{arXiv preprint arXiv:2109.07946}, 2021.

\bibitem{ciampaglia2018algorithmic}
G.~L. Ciampaglia, A.~Nematzadeh, F.~Menczer, and A.~Flammini, ``How algorithmic
  popularity bias hinders or promotes quality,'' \emph{Scientific reports},
  vol.~8, no.~1, pp. 1--7, 2018.

\bibitem{chaney2018algorithmic}
A.~J. Chaney, B.~M. Stewart, and B.~E. Engelhardt, ``How algorithmic
  confounding in recommendation systems increases homogeneity and decreases
  utility,'' in \emph{Proceedings of the 12th ACM conference on recommender
  systems}, 2018, pp. 224--232.

\bibitem{krauth2020offline}
K.~Krauth, S.~Dean, A.~Zhao, W.~Guo, M.~Curmei, B.~Recht, and M.~I. Jordan,
  ``Do offline metrics predict online performance in recommender systems?''
  \emph{arXiv preprint arXiv:2011.07931}, 2020.

\bibitem{lucherini2021t}
E.~Lucherini, M.~Sun, A.~Winecoff, and A.~Narayanan, ``T-recs: A simulation
  tool to study the societal impact of recommender systems,'' \emph{arXiv
  preprint arXiv:2107.08959}, 2021.

\bibitem{bountouridis2019siren}
D.~Bountouridis, J.~Harambam, M.~Makhortykh, M.~Marrero, N.~Tintarev, and
  C.~Hauff, ``Siren: A simulation framework for understanding the effects of
  recommender systems in online news environments,'' in \emph{Proceedings of
  the conference on fairness, accountability, and transparency}, 2019, pp.
  150--159.

\bibitem{yao2021measuring}
S.~Yao, Y.~Halpern, N.~Thain, X.~Wang, K.~Lee, F.~Prost, E.~H. Chi, J.~Chen,
  and A.~Beutel, ``Measuring recommender system effects with simulated users,''
  \emph{arXiv preprint arXiv:2101.04526}, 2021.

\bibitem{zhou2021longitudinal}
M.~Zhou, J.~Zhang, and G.~Adomavicius, ``Longitudinal impact of preference
  biases on recommender systems' performance,'' \emph{Kelley School of Business
  Research Paper}, no. 2021-10, 2021.

\bibitem{ma2018entire}
X.~Ma, L.~Zhao, G.~Huang, Z.~Wang, Z.~Hu, X.~Zhu, and K.~Gai, ``Entire space
  multi-task model: An effective approach for estimating post-click conversion
  rate,'' in \emph{The 41st International ACM SIGIR Conference on Research \&
  Development in Information Retrieval}, 2018, pp. 1137--1140.

\bibitem{saito2020unbiased}
Y.~Saito, ``Unbiased pairwise learning from biased implicit feedback,'' in
  \emph{Proceedings of the 2020 ACM SIGIR on International Conference on Theory
  of Information Retrieval}, 2020, pp. 5--12.

\bibitem{pennington2014glove}
J.~Pennington, R.~Socher, and C.~D. Manning, ``Glove: Global vectors for word
  representation,'' in \emph{Proceedings of the 2014 conference on empirical
  methods in natural language processing (EMNLP)}, 2014, pp. 1532--1543.

\bibitem{devlin2018bert}
J.~Devlin, M.-W. Chang, K.~Lee, and K.~Toutanova, ``Bert: Pre-training of deep
  bidirectional transformers for language understanding,'' \emph{arXiv preprint
  arXiv:1810.04805}, 2018.

\bibitem{allen2019analogies}
C.~Allen and T.~Hospedales, ``Analogies explained: Towards understanding word
  embeddings,'' in \emph{International Conference on Machine Learning}.\hskip
  1em plus 0.5em minus 0.4em\relax PMLR, 2019, pp. 223--231.

\bibitem{ethayarajh2019towards}
K.~Ethayarajh, D.~Duvenaud, and G.~Hirst, ``Towards understanding linear word
  analogies,'' in \emph{Proceedings of the 57th Annual Meeting of the
  Association for Computational Linguistics}, 2019, pp. 3253--3262.

\bibitem{del2017datagencars}
M.~del Carmen Rodr{\'\i}guez-Hern{\'a}ndez, S.~Ilarri, R.~Hermoso, and
  R.~Trillo-Lado, ``Datagencars: A generator of synthetic data for the
  evaluation of context-aware recommendation systems,'' \emph{Pervasive and
  Mobile Computing}, vol.~38, pp. 516--541, 2017.

\bibitem{jakomin2018generating}
M.~Jakomin, T.~Curk, and Z.~Bosni{\'c}, ``Generating inter-dependent data
  streams for recommender systems,'' \emph{Simulation Modelling Practice and
  Theory}, vol.~88, pp. 1--16, 2018.

\bibitem{slokom2018comparing}
M.~Slokom, ``Comparing recommender systems using synthetic data,'' in
  \emph{Proceedings of the 12th ACM Conference on Recommender Systems}, 2018,
  pp. 548--552.

\bibitem{rohde2018recogym}
D.~Rohde, S.~Bonner, T.~Dunlop, F.~Vasile, and A.~Karatzoglou, ``Recogym: A
  reinforcement learning environment for the problem of product recommendation
  in online advertising,'' \emph{arXiv preprint arXiv:1808.00720}, 2018.

\bibitem{chen2019generative}
X.~Chen, S.~Li, H.~Li, S.~Jiang, Y.~Qi, and L.~Song, ``Generative adversarial
  user model for reinforcement learning based recommendation system,'' in
  \emph{International Conference on Machine Learning}, 2019, pp. 1052--1061.

\bibitem{ie2019recsim}
E.~Ie, C.-w. Hsu, M.~Mladenov, V.~Jain, S.~Narvekar, J.~Wang, R.~Wu, and
  C.~Boutilier, ``Recsim: A configurable simulation platform for recommender
  systems,'' \emph{arXiv preprint arXiv:1909.04847}, 2019.

\bibitem{mladenov2021recsim}
M.~Mladenov, C.-W. Hsu, V.~Jain, E.~Ie, C.~Colby, N.~Mayoraz, H.~Pham, D.~Tran,
  I.~Vendrov, and C.~Boutilier, ``Recsim ng: Toward principled uncertainty
  modeling for recommender ecosystems,'' \emph{arXiv preprint
  arXiv:2103.08057}, 2021.

\bibitem{shi2019virtual}
J.-C. Shi, Y.~Yu, Q.~Da, S.-Y. Chen, and A.-X. Zeng, ``Virtual-taobao:
  Virtualizing real-world online retail environment for reinforcement
  learning,'' in \emph{Proceedings of the AAAI Conference on Artificial
  Intelligence}, vol.~33, no.~01, 2019, pp. 4902--4909.

\bibitem{huang2020keeping}
J.~Huang, H.~Oosterhuis, M.~De~Rijke, and H.~Van~Hoof, ``Keeping dataset biases
  out of the simulation: A debiased simulator for reinforcement learning based
  recommender systems,'' in \emph{Fourteenth ACM conference on recommender
  systems}, 2020, pp. 190--199.

\bibitem{mcinerney2021accordion}
J.~McInerney, E.~Elahi, J.~Basilico, Y.~Raimond, and T.~Jebara, ``Accordion: A
  trainable simulator for long-term interactive systems,'' in \emph{Fifteenth
  ACM Conference on Recommender Systems}, 2021, pp. 102--113.

\bibitem{zhao2021usersim}
X.~Zhao, L.~Xia, L.~Zou, H.~Liu, D.~Yin, and J.~Tang, ``Usersim: User
  simulation via supervised generativeadversarial network,'' in
  \emph{Proceedings of the Web Conference 2021}, 2021, pp. 3582--3589.

\bibitem{wang2021learning}
W.~Wang, ``Learning to recommend from sparse data via generative user
  feedback,'' in \emph{Proceedings of the AAAI Conference on Artificial
  Intelligence}, vol.~35, no.~5, 2021, pp. 4436--4444.

\bibitem{chen2020bias}
J.~Chen, H.~Dong, X.~Wang, F.~Feng, M.~Wang, and X.~He, ``Bias and debias in
  recommender system: A survey and future directions,'' \emph{arXiv preprint
  arXiv:2010.03240}, 2020.

\bibitem{imbens2015causal}
G.~W. Imbens and D.~B. Rubin, \emph{Causal inference in statistics, social, and
  biomedical sciences}.\hskip 1em plus 0.5em minus 0.4em\relax Cambridge
  University Press, 2015.

\bibitem{bishop2006pattern}
C.~M. Bishop and N.~M. Nasrabadi, \emph{Pattern recognition and machine
  learning}.\hskip 1em plus 0.5em minus 0.4em\relax Springer, 2006, vol.~4,
  no.~4.

\bibitem{rendle2012bpr}
S.~Rendle, C.~Freudenthaler, Z.~Gantner, and L.~Schmidt-Thieme, ``Bpr: Bayesian
  personalized ranking from implicit feedback,'' \emph{arXiv preprint
  arXiv:1205.2618}, 2012.

\bibitem{wilson2014humans}
R.~C. Wilson, A.~Geana, J.~M. White, E.~A. Ludvig, and J.~D. Cohen, ``Humans
  use directed and random exploration to solve the explore--exploit dilemma.''
  \emph{Journal of Experimental Psychology: General}, vol. 143, no.~6, p. 2074,
  2014.

\bibitem{gershman2018deconstructing}
S.~J. Gershman, ``Deconstructing the human algorithms for exploration,''
  \emph{Cognition}, vol. 173, pp. 34--42, 2018.

\bibitem{gulla2017adressa}
J.~A. Gulla, L.~Zhang, P.~Liu, {\"O}.~{\"O}zg{\"o}bek, and X.~Su, ``The adressa
  dataset for news recommendation,'' in \emph{Proceedings of the international
  conference on web intelligence}, 2017, pp. 1042--1048.

\bibitem{heinzerling2018bpemb}
B.~Heinzerling and M.~Strube, ``{BPEmb: Tokenization-free Pre-trained Subword
  Embeddings in 275 Languages},'' in \emph{Proceedings of the Eleventh
  International Conference on Language Resources and Evaluation (LREC 2018)},
  2018.

\bibitem{shannon1948mathematical}
C.~E. Shannon, ``A mathematical theory of communication,'' \emph{The Bell
  system technical journal}, vol.~27, no.~3, pp. 379--423, 1948.

\end{thebibliography}
